# Real-Time Proactive Anomaly Detection via Forward and Backward Forecast Modeling

Luis Olmos[1][a], Rashida Hasan[2][b]
[c]

[1,2]*Department of Computer Science, California State University, Northridge, California, USA*
[1]*luis.olmos.563@my.csun.edu*, [2]*rashida.hasan@csun.edu*

Abstract: Reactive anomaly detection methods, which are commonly deployed to identify anomalies after they occur based on observed deviations, often fall short in applications that demand timely intervention, such as industrial monitoring, finance, and cybersecurity. Proactive anomaly detection, by contrast, aims to detect early warning signals before failures fully manifest, but existing methods struggle with handling heterogeneous multivariate data and maintaining precision under noisy or unpredictable conditions. In this work, we introduce two proactive anomaly detection frameworks: the Forward Forecasting Model (FFM) and the Backward Reconstruction Model (BRM). Both models leverage a hybrid architecture combining Temporal Convolutional Networks (TCNs), Gated Recurrent Units (GRUs), and Transformer encoders to model directional temporal dynamics. FFM forecasts future sequences to anticipate disruptions, while BRM reconstructs recent history from future context to uncover early precursors. Anomalies are flagged based on forecasting error magnitudes and directional embedding discrepancies. Our models support both continuous and discrete multivariate features, enabling robust performance in real-world settings. Extensive experiments on four benchmark datasets, MSL, SMAP, SMD, and PSM, demonstrate that FFM and BRM outperform state-of-the-art baselines across detection metrics and significantly improve the timeliness of anomaly anticipation. These properties make our approach well-suited for deployment in time-sensitive domains requiring proactive monitoring.

## 1 INTRODUCTION

Time series anomaly detection plays a critical role in maintaining the safety, reliability, and performance of complex systems. The objective is to detect deviations from expected behavior in temporal data, signals that may correspond to system faults, intrusions, or impending failures. Traditional approaches are primarily reactive and frequently fall short in addressing the complex temporal dependencies, high-dimensional heterogeneity, and sparse labeling that characterize real-world time series. In this work, we use the term real-time to denote low-latency, online inference without access to future ground-truth observations, and proactive to describe detection based on forecasted values rather than observed data.

Reactive anomaly detection techniques are commonly applied in a retrospective setting, where anomalies are identified based on discrepancies between observed values and learned normal patterns. These approaches include (i) statistical forecasting-based models, such as ARIMA and Gaussian models, which detect anomalies via deviations from predicted trends; (ii) classical unsupervised detectors, such as k-nearest neighbors and density-based methods, applied to time-series representations; and (iii) deep learning-based models, including autoencoders, transformers [Tuli et al., 2022, Xu et al., 2021], and graph-based methods [Deng and Hooi, 2021]. While these methods are effective for post-hoc detection, they are most often deployed in reactive settings and typically flag anomalies only after abnormal behavior becomes observable.

To overcome these limitations, proactive anomaly detection has emerged as a promising paradigm. Proactive anomaly detection frameworks aim to forecast future states and identify deviations before ob-

---

servable anomalies manifest. Notable recent approaches include RePAD [Lee et al., 2020], which introduces real-time proactive detection by leveraging prediction error trends, and AutoGuard [Madi et al., 2021], which uses dual intelligence at the application layer in 5G networks to anticipate threats. Forecasting-based anomaly detection using deep autoencoders has also been applied in epidemiological settings [Rancati et al., 2024]. Despite this progress, many proactive approaches still rely on unidirectional forecasting, and are not real-time and not generalized, limiting their understanding of temporal symmetry and asymmetry. This motivates our novel approach, which integrates directional modeling to capture nuanced predictive discrepancies.

In this work, we introduce two distinct proactive anomaly detection frameworks: the Forward Forecasting Model (FFM) and the Backward Reconstruction Model (BRM). Each model offers a unique perspective on anticipating anomalies before they fully manifest. FFM (Forward Forecasting Model) is designed to predict future time series segments based on historical context. By leveraging a hybrid deep architecture consisting of Temporal Convolutional Networks (TCNs), Gated Recurrent Units (GRUs), and Transformer encoders, FFM captures both short- and long-term temporal patterns. Anomalies are flagged when the forecasting error, the deviation between predicted and observed future values, exceeds learned thresholds, indicating unexpected behavior on the horizon. BRM (Backward Reconstruction Model) takes a contrasting approach. It operates by reconstructing recent historical data using future observations, essentially modeling the sequence in reverse. Like FFM, BRM uses the same hybrid architecture (TCN–GRU–Transformer), but inverts the direction of analysis. Anomalies are detected when reconstruction fidelity deteriorates, signaling inconsistencies between anticipated and actual past behavior. Unlike traditional reactive models that only respond after an anomaly becomes visible, our proactive models aim to surface meaningful precursors that allow earlier and more informed intervention. Both FFM and BRM are designed to support heterogeneous multivariate inputs, including both continuous and discrete features, making them suitable for complex real-world environments such as industrial systems, financial platforms, and cyber-physical infrastructures.

We evaluate our approach not only against ground-truth anomaly labels but also in terms of its ability to forecast true future values during anomaly windows. This dual evaluation-against both anomaly detection metrics and predictive fidelity-offers a more rigorous assessment of a model's proactive capability.

We benchmark our framework on four widely-used multivariate time series datasets (MSL, SMAP, SMD, and PSM) [Hundman et al., 2018, Su et al., 2019, Abdulaal et al., 2021] and demonstrate superior performance over state-of-the-art baselines, both in detection accuracy and in the timeliness of anomaly anticipation. The key contributions of our work are:

- Two Proactive Models: We propose FFM and BRM, standalone models for forward forecasting and backward reconstruction, designed to anticipate anomalies before they fully emerge.

- Shared Hybrid Architecture: Both models employ a unified backbone combining TCN, GRU, and Transformer layers to capture multiscale temporal dependencies in noisy, high-dimensional data.

- Flexible Deployment: FFM supports online forecasting, while BRM enables hindsight-based detection, offering adaptability based on application latency constraints.

- Strong Empirical Results: Experiments on MSL, SMAP, SMD, and PSM show that FFM and BRM outperform state-of-the-art methods in accuracy and lead time across diverse domains.

**Reproducibility**: All of our code and datasets are available on GitHub Repository

## 2 RELATED WORK

Anomaly detection in time series has evolved from reactive to proactive paradigms. Broadly, existing approaches can be categorized into two types: reactive methods, which identify anomalies at the time of occurrence, and proactive methods, which aim to anticipate anomalies in advance using forecasting techniques.

Reactive methods have been extensively studied. The Anomaly Transformer [Xu et al., 2021] uses self-attention to detect local-global discrepancies but incurs high complexity and lacks foresight. TranAD [Tuli et al., 2022] boosts accuracy via adversarial and self-conditioned transformers, yet remains reactive and unsuitable for streaming.

Other unsupervised time series anomaly detection methods include LSTM-P [Malhotra et al., 2015], which employs a two-layer stacked LSTM followed by a fully connected output layer for sequence prediction. DeepAnT [Munir et al., 2018] adopts a CNN-based architecture with stacked 1D convolution and max pooling layers for temporal modeling. TCN-S2S-P [He and Zhao, 2019] utilizes temporal convolutional networks with dilated causal convolutions to capture long-range dependencies. GTA [Chen et al., 2021] extends transformer-based modeling by learning latent graph structures and encoding both temporal and relational dynamics. While effective in modeling complex temporal patterns, these methods primarily focus on reactive detection and often lack proactive capabilities.

In contrast, proactive methods focus on forecasting anomalies before they manifest. RePAD [Lee et al., 2020] exemplifies this with an online LSTM and adaptive thresholding to enable early warnings without pretraining. However, it is limited to univariate settings and is sensitive to initialization noise. To extend proactivity to multivariate time series, Jeon et al. [Jeon et al., 2025] propose a forecasting model with Adaptive Graph Convolution and Fourier-based analysis, showing strong performance on benchmark datasets like SMAP and MSL, though with reduced effectiveness in less predictable domains like SMD.

Despite growing interest, proactive anomaly detection remains largely underdeveloped, with only a handful of methods proposed to date. Moreover, these approaches often rely on application-specific assumptions or use forecasted values as ground truth, hindering generalizability and reproducible evaluation. To overcome the limitations of reactive methods and the limited progress in proactive approaches, we introduce two predictive models: the Forward Forecasting Model (FFM) and the Backward Reconstruction Model (BRM). Unlike reactive systems, our models operate purely in anticipatory mode, enabling timely detection across diverse scenarios.

We propose two proactive anomaly detection frameworks with a shared architecture: the Forward Forecasting Model (FFM) and the Backward Reconstruction Model (BRM). Both models learn temporal dynamics and identify anomalies by detecting regions where predictions fail, indicating disruptions in the data's temporal structure. FFM predicts future time windows from historical inputs, while BRM reconstructs past sequences from future context. These prediction or reconstruction errors serve as early indicators of abnormal behavior. As illustrated in Figure 1, both models utilize a hybrid encoder composed of Temporal Convolutional Networks (TCN), Gated Recurrent Units (GRU), and Transformer layers. They jointly process discrete and continuous features, producing independent anomaly scores that reflect deviations from expected temporal patterns.

# 3 PROACTIVE ANOMALY DETECTION

We introduce two directional forecasting frameworks with a shared architecture: the Forward Forecasting Model (FFM) and the Backward Reconstruction Model (BRM). Both learn temporal dynamics and identify anomalies as regions where predictions fail, indicating disruptions in temporal structure. We define proactive detection as issuing a warning from an anomaly-free input window that forecasts an upcoming onset within the prediction horizon. Concretely, FFM predicts future windows from past inputs; BRM reconstructs past windows from future context. As illustrated in Figure 1, both models use a hybrid encoder (TCN, GRU, Transformer) and produce anomaly scores over jointly modeled discrete and continuous features.

## 3.1 Problem Setup

Let $\mathbf{X} = [\mathbf{x}_1, \ldots, \mathbf{x}_T] \in \mathbb{R}^{T \times D}$ denote a multivariate time series with $D$ features and $T$ time steps, where each feature vector $\mathbf{x}_t \in \mathbb{R}^D$ consists of both continuous and discrete elements. Specifically, we define $D_{\text{cont}}$ as the subset of continuous-valued features and $D_{\text{disc}}$ as the subset of discrete or binary-valued features. Given a window length $W$ and a forecast horizon $H$, our objective is to detect potential anomalies at time $t$ by applying a sequence-to-sequence prediction model.

- **Forward model:** predict $\mathbf{Y}_t = [\mathbf{x}_{t+1}, \ldots, \mathbf{x}_{t+H}]$ from $\mathbf{X}_{t-W+1:t}$.
- **Backward model:** reconstruct $\mathbf{X}_{t-W+1:t}$ from $\mathbf{X}_{t+1:t+H}$.

The core idea is that anomalies disrupt temporal patterns, leading to poor prediction or reconstruction accuracy, whether forecasting forward or reconstructing backward.

**Reactive anomaly detection.** Given a past window $\mathbf{X}_{t-W+1:t}$, a predictive model outputs $\hat{\mathbf{x}}_t$. A reactive method declares an anomaly at time $t$ if the discrepancy between the observation $\mathbf{x}_t$ and the prediction $\hat{\mathbf{x}}_t$ exceeds a threshold, i.e.,

$$\text{Anomaly at } t \iff d(\mathbf{x}_t, \hat{\mathbf{x}}_t) > \tau,$$

where $d(\cdot, \cdot)$ is a distance or error function and $\tau$ is an anomaly threshold. Thus, reactive detection requires access to the observed $\mathbf{x}_t$ and can only flag anomalies *after* they have occurred.

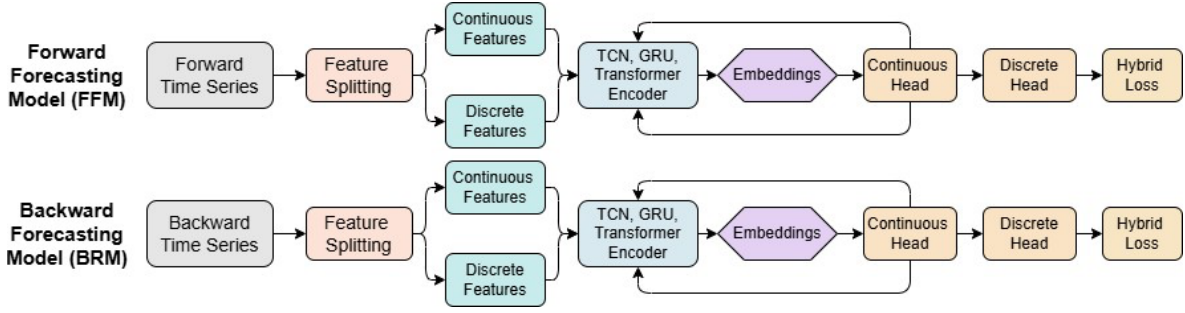

Figure 1: Overview of the proposed proactive anomaly detection framework. Arrows indicate sequential data flow from TCN to GRU to Transformer layers, followed by pooling and task-specific decoding for anomaly scoring.

**Proactive anomaly detection.** Given the same past window $\mathbf{X}_{t-W+1:t}$, a predictive model outputs a horizon of future predictions $\hat{\mathbf{Y}}_t = [\hat{\mathbf{x}}_{t+1}, \ldots, \hat{\mathbf{x}}_{t+H}]$. A proactive method declares an anomaly for time $t + h$ ($1 \leq h \leq H$) if the predicted value $\hat{\mathbf{x}}_{t+h}$ is deemed unlikely under the distribution of normal training data, i.e.,

$$\text{Anomaly at } t + h \iff f(\hat{\mathbf{x}}_{t+h}; D_{\text{train}}) < \tau,$$

where $f(\cdot; D_{\text{train}})$ is a data-driven normality score learned from training data and $\tau$ is a threshold. In this way, anomalies are identified based solely on forecasts, before $\mathbf{x}_{t+h}$ is observed, enabling earlier warning compared to reactive methods.

We follow Jeon et al. [Jeon et al., 2025] in using the term *proactive anomaly detection* to describe methods that detect anomalies from predicted values before ground-truth observations are available. Although our study fixes the horizon to $H = 1$ for consistency with prior work, the formulation is general for $H \geq 1$, enabling earlier detection when anomalies exhibit precursors. In practice, sliding windows may overlap with anomalous segments once an anomaly has already started; this limitation is inherent to forecasting-based detection, but the proactive setting still provides earlier warning compared to purely reactive approaches.

### 3.2 Forward Forecasting Model (FFM)

Anomalies degrade the predictability of future values by introducing irregularities that distort underlying patterns and violate model assumptions. By training a model to accurately forecast the immediate future using a past window, we can detect deviations in prediction quality as signals of abnormality. To do this effectively, we integrate: (i) a TCN for short/medium-term local patterns with dilated convolutions, (ii) a GRU for long-range memory in sequential dynamics,

(iii) a Transformer for global timepoint dependencies via attention, and (iv) a hybrid loss to jointly supervise continuous and discrete outputs. This layered combination allows the model to model rich temporal dynamics without sacrificing scalability or training efficiency.

To operationalize this architecture, our model processes an input window $\mathbf{X}_{t-W+1:t} \in \mathbb{R}^{W \times D}$, where $W$ is the window size and $D$ is the number of features. It passes through a sequence of temporal encoders, summarized below.

**Local Pattern Extraction with Temporal Convolution** We first use a Temporal Convolutional Network (TCN) to extract short- and medium-range temporal dependencies:

$$\mathbf{H}_{\text{TCN}} = \text{TCN}(\mathbf{X}_{t-W+1:t}) \in \mathbb{R}^{W \times C}. \quad (1)$$

$\mathbf{H}_{\text{TCN}}$ denotes the temporal feature maps produced by the TCN, and $C$ is the number of output channels. The use of dilated convolutions allows exponential receptive field growth, making it efficient for capturing repeating patterns and local trends.

**Long-Term Memory with GRU** To capture longer-range sequential dependencies beyond the convolutional window, we use a Gated Recurrent Unit (GRU):

$$\mathbf{H}_{\text{GRU}} = \text{GRU}(\mathbf{H}_{\text{TCN}}) \in \mathbb{R}^{W \times H_{\text{GRU}}}. \quad (2)$$

$\mathbf{H}_{\text{GRU}}$ contains the GRU-encoded hidden states, and $H_{\text{GRU}}$ is the GRU's hidden dimension. This component enhances the model's ability to capture temporal continuity and context.

**Global Dependency Modeling with Transformer** To refine the temporal features globally, we apply a

Transformer encoder, which uses self-attention to assess interactions between all timepoints:

$$Z_{enc} = \text{TransformerEncoder}(H_{GRU}) \in \mathbb{R}^{W \times H_{GRU}}. \quad (3)$$

$Z_{enc}$ contains the attention-refined features for each timestep. This step allows the model to adaptively focus on informative parts of the sequence, especially useful for multivariate and noisy inputs.

**Latent Representation via Mean Pooling** To produce a fixed-size latent representation from the temporal sequence, we perform mean pooling over the attention-encoded vectors:

$$z = \frac{1}{W} \sum_{i=1}^{W} Z_{enc}[i] \in \mathbb{R}^{H_{GRU}}. \quad (4)$$

The resulting vector $z$ serves as a compact summary of the temporal input window, preserving both local and global dynamics.

**Parallel Forecasting Heads** Finally, $z$ is passed to two separate linear heads for predicting continuous and discrete features over the future horizon $H$:

$$\hat{Y}^{cont} = \text{Linear}_{cont}(z) \in \mathbb{R}^{H \times |D_{cont}|},$$
$$\hat{Y}^{disc} = \text{Linear}_{disc}(z) \in \mathbb{R}^{H \times |D_{disc}|}. \quad (5)$$

These two heads, respectively, generate: (i) $\hat{Y}^{cont}$, continuous-valued predictions optimized using Huber loss, and (ii) $\hat{Y}^{disc}$, discrete (binary) logits trained with binary cross-entropy (BCE) loss. This architecture allows the model to handle heterogeneous time series with mixed feature types and to anticipate anomalies by capturing failures in forecasting either feature type. To jointly supervise heterogeneous outputs, we use a hybrid loss:

$$L_{\rightarrow} = \alpha \cdot L_{Huber} + \beta \cdot L_{BCE}, \quad (6)$$

Here, $L_{Huber}$ computes the smoothed regression error across all continuous-valued features and time steps, and $L_{BCE}$ applies binary cross-entropy loss to discrete outputs. The weights $\alpha$ and $\beta$ control the contribution of each term and are typically set to 1 but can be tuned for imbalanced or domain-specific data. This formulation is necessary for handling real-world multivariate time series, which often contain heterogeneous feature types, such as numeric sensors and binary flags. Using a unified loss would be inappropriate, as it would treat both types identically, leading to poor convergence or bias.

### 3.2.1 Anomaly Score Calculation

We explore two complementary strategies to derive anomaly predictions based on the model's forecast errors: one using an external unsupervised outlier detector, and the other using a threshold-free Top-$K$ ranking approach. In the first approach, we apply a post-hoc unsupervised anomaly detector to the model's prediction vectors. Specifically, we flatten the predicted future windows $\hat{Y}_t \in \mathbb{R}^{H \times D}$ into vectors $\hat{Y}_t^{flat} \in \mathbb{R}^{H \cdot D}$. An anomaly detection model (e.g., ECOD) is then trained on the training set's predictions and used to assign anomaly scores to test predictions. Anomalies are flagged by thresholding $s_t$ based on evaluation objectives.

In the second approach, we compute the mean squared forecast error between the predicted and true future values for each window:

$$s_t = \frac{1}{H \cdot D} \sum_{h=1}^{H} \sum_{d=1}^{D} ({}_t\hat{x}_{h,d} - {}_t x_{h,d})^2 \quad (7)$$

To mitigate the effect of extreme values and stabilize the score distribution, we apply a log-transform:

$$s_t \leftarrow \log(1 + s_t) \quad (8)$$

Assuming the number of ground truth anomalies $K$ is known, we sort the scores $\{s_t\}_{t=1}^{T'}$ and flag the top-$K$ highest scoring windows as anomalies:

$$\hat{a_t} = \begin{cases} 1 & \text{if } s_t \in \text{TopK}(\{s_i\}_{i=1}^{T'}) \\ 0 & \text{otherwise} \end{cases} \quad (9)$$

This method avoids threshold tuning and is commonly adopted in real-world unsupervised evaluation settings. Algorithm 1 outlines the detailed procedure of the Forward Forecasting Model (FFM).

---
**Algorithm 1: Forward Forecasting Detection**

**Input:** time series $X$, window size $W$, forecast horizon $H$
**for** $t \leftarrow W$ **to** $T - H$ **do**
  $X_{input} \leftarrow X_{t-W+1:t}$
  $\hat{Y}_t \leftarrow F_{\rightarrow}(X_{input})$
  $\vec{s_t} \leftarrow \|\hat{Y}_t - X_{t+1:t+H}\|_1$
**end**
**return** anomaly scores $\{\vec{s_t}\}$

---

## 3.3 Backward Reconstruction Model (BRM)

The backward model $F_{\leftarrow}$ is designed to reconstruct a past input window $X_{t-W+1:t} \in \mathbb{R}^{W \times D}$ using only fu-ture observations $\mathbf{X}_{t+1:t+H} \in \mathbb{R}^{H \times D}$. The underlying premise is that anomalies disturb the temporal continuity not just forward in time, but also backward: future context often reflects abnormal signals that can be used to reconstruct what should have occurred in the recent past. Thus, BRM enables proactive anomaly detection by reasoning "in reverse." The model architecture mirrors the forward model, with temporal direction reversed.

- **Temporal Convolution (TCN):** Extracts local temporal patterns from the future window via dilated convolutions:

$$\mathbf{H}^{\leftarrow}_{\text{TCN}} = \text{TCN}(\mathbf{X}_{t+1:t+H}) \in \mathbb{R}^{H \times C} \quad (10)$$

where $C$ is the number of output channels.

- **Sequential Encoding (GRU):** Captures longer-range dependencies across the convolved future representations:

$$\mathbf{H}^{\leftarrow}_{\text{GRU}} = \text{GRU}(\mathbf{H}^{\leftarrow}_{\text{TCN}}) \in \mathbb{R}^{H \times H_{\text{GRU}}} \quad (11)$$

where $H_{\text{GRU}}$ is the GRU hidden size.

- **Global Attention (Transformer Encoder):** Refines temporal embeddings through self-attention across the entire future sequence:

$$\mathbf{Z}^{\leftarrow}_{\text{enc}} = \text{TransformerEncoder}(\mathbf{H}^{\leftarrow}_{\text{GRU}}) \in \mathbb{R}^{H \times H_{\text{GRU}}} \quad (12)$$

- **Latent Pooling:** Averages the time-dimension embeddings to obtain a fixed-length latent vector:

$$\mathbf{z}^{\leftarrow} = \frac{1}{H} \sum_{i=1}^{H} \mathbf{Z}^{\leftarrow}_{\text{enc}}[i] \in \mathbb{R}^{H_{\text{GRU}}} \quad (13)$$

- **Decoder:** Uses two linear heads to reconstruct both continuous and discrete components of the past sequence:

$$\hat{\mathbf{X}}_{t-W+1:t} = \text{Decoder}(\mathbf{z}^{\leftarrow}) \in \mathbb{R}^{W \times D} \quad (14)$$

This structure allows BRM to predict past behavior using only future data, exploiting the symmetric temporal dependencies often present in real-world systems.

### 3.3.1 Loss Function

The model is trained using the same hybrid loss as the forward model. The hybrid formulation ensures that the model remains sensitive to both numerical deviations and categorical transitions in the historical data.

### 3.3.2 Anomaly Score Computation

Anomalies are detected by comparing the reconstructed past $\hat{\mathbf{X}}_{t-W+1:t}$ to the actual past window $\mathbf{X}_{t-W+1:t}$ using mean squared error:

$$s^{\leftarrow}_t = \frac{1}{W \cdot D} \sum_{w=1}^{W} \sum_{d=1}^{D} (\hat{x}_{t-w+1,d} - x_{t-w+1,d})^2 \quad (15)$$

We apply a logarithmic transformation:

$$s^{\leftarrow}_t \leftarrow \log(1 + s^{\leftarrow}_t) \quad (16)$$

This scoring strategy quantifies how well future information can "explain" the past. Large reconstruction errors suggest that the future context is inconsistent with normal historical behavior which indicates a potential anomaly.

---

**Algorithm 2: Backward Forecasting Detection**

**Input:** time series $\mathbf{X}$, window size $W$, forecast horizon $H$
**for** $t \leftarrow W$ **to** $T - H$ **do**
    $\mathbf{X}_{\text{future}} \leftarrow \mathbf{X}_{t+1:t+H}$
    $\hat{\mathbf{X}}_{t-W+1:t} \leftarrow F_{\leftarrow}(\mathbf{X}_{\text{future}})$
    $s^{\leftarrow}_t \leftarrow \|\hat{\mathbf{X}}_{t-W+1:t} - \mathbf{X}_{t-W+1:t}\|_1$
**end**
**return** anomaly scores $\{s^{\leftarrow}_t\}$

---

## 3.4 Complexity Analysis

We analyze the computational complexity of our proposed forward and backward forecasting models in terms of time, highlighting their suitability for real-time and streaming anomaly detection scenarios. To facilitate the complexity analysis of our proposed algorithm, we first introduce the following notations: Let $B$ denote the batch size (treated as a constant in asymptotic analysis), $W$ the input window length, $H$ the forecast horizon, $D$ the number of input features, $C$ the number of TCN output channels, $H_{\text{GRU}}$ the hidden size of the GRU and Transformer, number of TCN layers $L_{\text{TCN}}$, and $L$ the number of Transformer layers (typically $L = 1$).

The total time complexity per forward pass is:

$$O\left(L_{\text{TCN}} \cdot BCDW + BWCH_{\text{GRU}} \\ + BW^2 H_{\text{GRU}} + BH_{\text{GRU}} D\right) \quad (17)$$

This includes: TCN ($L_{\text{TCN}}$ layers of dilated convolutions over $W$ time steps), GRU (sequential processing over $W$ steps), Transformer (self-attention contributes $O(W^2)$, which dominates for larger $W$), and Output Heads (final MLP layers are negligible).

**Inference Efficiency:** Backpropagation activations are not stored during inference. Additionally, small window sizes (e.g., $W = 5$–$20$) and a batch size of $B = 1$ are commonly used. These constraints help keep memory usage and latency low, enabling practical deployment for real-time and streaming anomaly detection.

## 4 EXPERIMENTS AND RESULT ANALYSIS

We evaluate our proactive anomaly detection model on four benchmark datasets, MSL, SMAP, SMD, and PSM, against traditional and state-of-the-art baselines. Specifically, we assess whether it (i) consistently outperforms time-series reactive methods and forecasting models in different performance metrics, (ii) leverages its proactive design to surpass retrospective models, (iii) improves detection using forecasted values as pseudo ground truth, and (iv) achieves strong performance when validated with true anomaly labels.

### 4.1 Dataset

We evaluate our model on four widely used multivariate time series datasets. MSL and SMAP [Hundman et al., 2018], released by NASA, contain labeled telemetry data from the Curiosity Rover and an Earth-observing satellite, respectively. Anomalies in these datasets stem from hardware faults and environmental disturbances. SMD [Su et al., 2019] consists of metrics from 28 servers monitored over several months, capturing diverse failure modes such as CPU/memory overloads and service outages. PSM [Abdulaal et al., 2021] provides sensor readings from power grid substations, with anomalies due to maintenance, power surges, and grid faults. For each dataset, we follow the original split between training (normal-only data) and testing (including both normal and anomalous sequences). A detailed summary of the training/testing splits and anomaly ratios is provided in Table 1.

Table 1: Dataset Statistics Used in Experiments

| Dataset | Train | Test | Anomaly Ratio (%) |
|---|---|---|---|
| MSL | 46,655 | 85,391 | 10.5 |
| SMAP | 108,148 | 454,652 | 12.8 |
| SMD | 566,725 | 850,100 | 4.16 |
| PSM | 103,289 | 114,336 | 27.76 |

### 4.2 Experimental Configuration

Each model is trained independently using the Adam optimizer with a learning rate of $10^{-4}$ and a weight decay of $10^{-5}$ for 5 epochs. We apply gradient clipping to a maximum norm of 1.0 to prevent exploding gradients and include a dropout layer before the decoder to regularize training.

Following prior work [Jeon et al., 2025], we adopt a fixed input window of length $W = 5$ and a forecast horizon of $H = 1$. At each time step $t$, the model observes the past window $\mathbf{X}_{t-W+1:t}$ and predicts the next value $\hat{\mathbf{x}}_{t+1}$.

This setup provides a form of *proactive detection*: anomalies can be flagged based on predicted values before the ground-truth $\mathbf{x}_{t+1}$ is observed. In practice, because sliding windows advance step by step, some input windows may overlap with anomalous regions once an anomaly has already begun. This overlap is unavoidable in forecasting-based evaluation and has also been reported in prior work. Nonetheless, when anomalies exhibit precursors, our method can forecast abnormal values ahead of their occurrence, thus providing earlier warning than reactive error-based methods. We emphasize that our horizon is limited to $H = 1$ for consistency and comparability with prior studies, but the framework is naturally extensible to longer horizons. For each dataset, we follow the original split between training (normal-only data) and testing (including both normal and anomalous sequences).

To ensure a fair comparison with state-of-the-art methods, we use the same three anomaly detectors as part of our evaluation framework, following the approach in [Jeon et al., 2025]. Specifically, we employ:

1. Gaussian Mixture Model (GMM), a probabilistic clustering technique that computes anomaly scores based on the likelihood of data points under learned Gaussian components.

2. Empirical Cumulative Outlier Detection (ECOD) [Li et al., 2022], a non-parametric and non-deep anomaly detection method that estimates empirical outlier probabilities using feature-wise statistics.

3. DeepSVDD [Schölkopf et al., 1999], a deep one-class classification method that maps data into a hypersphere and identifies anomalies based on their distance from the center.

To evaluate detection performance, we adopt four metrics commonly used in recent anomaly detection studies. F1-@K [Kim et al., 2022] computes the F1 score over the top-$K$ highest-scoring windows, where $K$ is the number of true anomalies, favoring early and high-confidence detections. F1-C (Composite F1) [Garg et al., 2022] considers any overlap with the ground truth anomaly range as a true positive, emphasizing coverage. F1-R (Range-Based F1) [Wagner et al., 2023] incorporates both detection accuracy and temporal alignment within anomalous segments. Finally, F1 (Overall) captures the standard precision-recall balance across all predicted windows. The bold values in all tables indicate superior performance by our algorithm. All experiments are conducted on a single NVIDIA T4 GPU to ensure consistency and reproducibility.

Table 2: Comparison with Time-Series Anomaly Detection and Forecasting Models on MSL Dataset

| Data-driven Model | GMM | | | ECOD | | | DeepSVDD | | |
|---|---|---|---|---|---|---|---|---|---|
| | F1-@K | F1-C | F1-R | F1-@K | F1-C | F1-R | F1-@K | F1-C | F1-R |
| LSTM-P | 0.1906 | 0.1906 | 0.1905 | 0.1906 | 0.1906 | 0.1905 | 0.2026 | 0.1982 | 0.1973 |
| DeepAnT | 0.0451 | 0.2179 | 0.0404 | 0.1906 | 0.1906 | 0.1905 | 0.1906 | 0.1906 | 0.1905 |
| TCN-S2S-P | 0.1906 | 0.1906 | 0.1905 | 0.1906 | 0.1906 | 0.1905 | 0.1908 | 0.1908 | 0.1907 |
| GTA | 0.1906 | 0.1906 | 0.1905 | 0.1906 | 0.1906 | 0.1905 | 0.1913 | 0.1910 | 0.1908 |
| DLinear | 0.1906 | 0.1906 | 0.1905 | 0.1906 | 0.1906 | 0.1905 | **0.2213** | 0.2169 | **0.2142** |
| PatchTST | **0.1906** | 0.1906 | **0.1905** | 0.1906 | 0.1906 | **0.1905** | 0.2014 | 0.1996 | 0.1987 |
| FFM (ours) | 0.1171 | **0.3169** | 0.1092 | 0.1228 | **0.2403** | 0.1026 | 0.0471 | **0.2250** | 0.0411 |
| BRM (ours) | 0.1602 | 0.2303 | 0.1402 | **0.2001** | 0.1201 | 0.1402 | 0.1801 | 0.1750 | 0.1401 |

Table 3: Comparison with Time-Series Anomaly Detection and Forecasting Models on SMAP Dataset

| Data-driven Model | GMM | | | ECOD | | | DeepSVDD | | |
|---|---|---|---|---|---|---|---|---|---|
| | F1-@K | F1-C | F1-R | F1-@K | F1-C | F1-R | F1-@K | F1-C | F1-R |
| LSTM-P | 0.2269 | 0.2268 | 0.2268 | 0.2031 | 0.1792 | 0.1491 | 0.2501 | 0.2393 | 0.2332 |
| DeepAnT | 0.0001 | 0.0059 | 0.0000 | 0.1981 | **0.2040** | 0.1669 | 0.2268 | 0.2268 | 0.2268 |
| TCN-S2S-P | 0.2268 | 0.2268 | 0.2268 | 0.1284 | 0.1464 | 0.0918 | 0.2528 | 0.2438 | 0.2374 |
| GTA | **0.2277** | 0.2268 | **0.2269** | 0.2348 | 0.2020 | **0.1731** | **0.2717** | **0.2540** | **0.2444** |
| DLinear | 0.2268 | 0.2268 | 0.2268 | 0.1990 | 0.0918 | 0.0194 | 0.2309 | 0.2254 | 0.2225 |
| PatchTST | 0.2268 | 0.2268 | 0.2268 | 0.2021 | 0.0887 | 0.0194 | 0.2282 | 0.2234 | 0.2209 |
| FFM (ours) | 0.1901 | 0.2005 | 0.1202 | 0.0591 | 0.1101 | 0.0150 | 0.0950 | 0.0820 | 0.0750 |
| BRM (ours) | 0.2101 | **0.2505** | 0.1408 | 0.0350 | 0.0850 | 0.0104 | 0.0910 | 0.0740 | 0.0820 |

Table 4: Comparison with Time-Series Anomaly Detection and Forecasting Models on SMD Dataset

| Data-driven Model | GMM | | | ECOD | | | DeepSVDD | | |
|---|---|---|---|---|---|---|---|---|---|
| | F1-@K | F1-C | F1-R | F1-@K | F1-C | F1-R | F1-@K | F1-C | F1-R |
| LSTM-P | 0.1240 | 0.1077 | 0.1050 | **0.1468** | **0.2125** | **0.0636** | 0.0920 | 0.0763 | 0.0757 |
| DeepAnT | 0.1131 | 0.0979 | 0.0554 | 0.0424 | 0.0637 | 0.0124 | 0.0928 | 0.0754 | 0.0746 |
| TCN-S2S-P | 0.0832 | 0.0811 | 0.0809 | 0.0203 | 0.0612 | 0.0033 | 0.0924 | 0.0766 | 0.0756 |
| GTA | 0.1047 | 0.0845 | 0.0768 | 0.0006 | 0.0011 | 0.0001 | 0.0901 | 0.0757 | 0.0749 |
| DLinear | 0.0917 | 0.1062 | 0.0494 | 0.0358 | 0.0660 | 0.0103 | 0.0917 | 0.0785 | 0.0778 |
| PatchTST | 0.1190 | 0.1489 | 0.0699 | 0.0410 | 0.0757 | 0.0121 | 0.0910 | 0.0773 | 0.0767 |
| FFM (ours) | 0.1901 | 0.2005 | 0.1202 | 0.0591 | 0.1101 | 0.0150 | **0.0950** | **0.0820** | 0.0750 |
| BRM (ours) | **0.2101** | **0.2505** | **0.1408** | 0.0350 | 0.0850 | 0.0104 | 0.0910 | 0.0740 | **0.0820** |

Table 5: Comparison with Time-Series Anomaly Detection and Forecasting Models on PSM Dataset

| Data-driven Model | GMM | | | ECOD | | | DeepSVDD | | |
|---|---|---|---|---|---|---|---|---|---|
| | F1-@K | F1-C | F1-R | F1-@K | F1-C | F1-R | F1-@K | F1-C | F1-R |
| LSTM-P | 0.0034 | 0.0160 | 0.0007 | 0.0016 | 0.0219 | 0.0013 | 0.4571 | 0.4347 | 0.4304 |
| DeepAnT | 0.0000 | 0.0000 | 0.0000 | 0.0087 | 0.0590 | 0.0057 | 0.4572 | 0.4358 | 0.4316 |
| TCN-S2S-P | 0.0000 | 0.0000 | 0.0000 | 0.0000 | 0.0000 | 0.0000 | 0.4505 | 0.4307 | 0.4241 |
| GTA | 0.0000 | 0.0000 | 0.0000 | 0.0005 | 0.0055 | 0.0002 | 0.4631 | 0.4428 | **0.4357** |
| DLinear | 0.0181 | 0.0540 | 0.0010 | 0.0239 | 0.0800 | 0.0069 | 0.4459 | 0.4331 | 0.4312 |
| PatchTST | 0.0244 | 0.0274 | 0.0006 | 0.0241 | 0.0800 | 0.0070 | 0.4459 | 0.4346 | 0.4326 |
| FFM (ours) | 0.0506 | 0.1304 | 0.0220 | 0.0305 | 0.1030 | 0.0104 | **0.4604** | **0.4404** | 0.4305 |
| BRM (ours) | **0.0610** | **0.1403** | **0.0302** | **0.0405** | **0.1101** | **0.0209** | 0.4505 | 0.4307 | 0.4208 |

Table 6: Evaluation Using Forecasted Values as Pseudo Ground Truth on MSL Dataset

| Proactive Anomaly Framework | GMM | | | | ECOD | | | | DeepSVDD | | | |
|---|---|---|---|---|---|---|---|---|---|---|---|---|
| | F1-@K | F1-C | F1-R | F1 | F1-@K | F1-C | F1-R | F1 | F1-@K | F1-C | F1-R | F1 |
| FFM (ours) | 0.1171 | 0.3169 | 0.1092 | 0.1303 | 0.1228 | **0.2403** | 0.1026 | 0.1601 | 0.0471 | **0.2250** | 0.0411 | 0.1501 |
| BRM (ours) | **0.1602** | 0.2303 | **0.1402** | 0.1202 | 0.2001 | 0.1201 | 0.1402 | 0.1900 | 0.1801 | 0.1750 | 0.1401 | 0.1650 |
| Proactive Forecasting | 0.0837 | 0.1972 | 0.0853 | 0.1801 | **0.2279** | 0.1940 | **0.1774** | **0.2201** | **0.1902** | 0.1870 | **0.1869** | **0.2201** |
| Anomaly Detector Truth | 0.0916 | 0.3428 | 0.0937 | 0.3001 | 0.0143 | 0.1502 | 0.0024 | 0.1104 | 0.1730 | 0.1471 | 0.1411 | 0.1401 |

Table 7: Evaluation Using Forecasted Values as Pseudo Ground Truth on SMAP Dataset

| Proactive Anomaly Framework | GMM | | | | ECOD | | | | DeepSVDD | | | |
|---|---|---|---|---|---|---|---|---|---|---|---|---|
| | F1-@K | F1-C | F1-R | F1 | F1-@K | F1-C | F1-R | F1 | F1-@K | F1-C | F1-R | F1 |
| FFM (ours) | 0.1501 | 0.1503 | **0.2505** | 0.2501 | 0.1107 | 0.2650 | 0.1413 | **0.2605** | 0.1302 | 0.2104 | 0.2002 | 0.2202 |
| BRM (ours) | **0.2540** | **0.3609** | 0.2301 | **0.2604** | **0.1810** | **0.2803** | **0.1503** | 0.2604 | **0.2604** | **0.3401** | **0.2502** | 0.2310 |
| Proactive Forecasting | 0.0185 | 0.2181 | 0.0026 | 0.1505 | 0.0210 | 0.1650 | 0.0014 | 0.2400 | 0.2416 | 0.2353 | 0.2314 | 0.2369 |
| Anomaly Detector Truth | 0.0367 | 0.3185 | 0.0078 | 0.1210 | 0.0290 | 0.2908 | 0.0046 | 0.2601 | 0.2521 | 0.2427 | 0.2364 | 0.2440 |

## 4.3 Comparison with Time-Series Anomaly Detection and Forecasting Models

We evaluate our proactive models, FFM and BRM, against four unsupervised anomaly detection methods: LSTM-P [Malhotra et al., 2015], DeepAnT [Munir et al., 2018], TCN-S2S-P [He and Zhao, 2019], GTA [Chen et al., 2021], and two forecasting-based approaches: DLinear [Zeng et al., 2023] and PatchTST [Nie et al., 2023]. LSTM-P, TCN-S2S-P, and DeepAnT are deep learning-based forecasting models using LSTM, CNN, and TCN architectures, respectively. GTA leverages a transformer with graph-based temporal modeling, while DLinear and PatchTST are recent forecasting models. Evaluation is performed on four standard datasets using GMM, ECOD, and DeepSVDD scoring methods (Tables 2–5).

In Tables 2–5, we observed cases where different models report identical performance scores. This arises when the models degenerate to predicting all samples as anomalous, which leads to a constant F1 baseline (e.g., 0.1906 for MSL or 0.2268 for SMAP). Such behavior was also noted in Jeon et al. [Jeon et al., 2025] and reflects a limitation of forecasting-based anomaly detection: if a model fails to learn normal patterns, the downstream anomaly detector will trivially classify the entire sequence as abnormal.

Across four datasets and three scoring methods, FFM and BRM continue to demonstrate strong performance relative to reactive baselines. There are 12 evaluations for each metric (F1-C, F1@K, F1-R), derived from 4 datasets and 3 anomaly detectors (GMM, ECOD, DeepSVDD) per dataset. Based on the F1-C results, FFM achieves the highest score in 3 cases, while BRM leads in 4, consistently outperforming baseline models such as DeepAnT, PatchTST, and GTA.

On MSL, FFM achieves the highest F1-C under GMM (0.3169) and ECOD (0.2403), while BRM is competitive under GMM (0.2303). On SMAP and SMD, BRM leads with top F1-C scores under GMM (0.2505), and FFM performs well under DeepSVDD. For PSM, FFM records the best F1-C (0.4404, DeepSVDD), with BRM close behind across other evaluations. Beyond classification performance, we observe complementary strengths in ranking-based metrics. Out of 12 evaluations, FFM leads F1-@K in 2 cases, BRM in 3. BRM leads F1-R in 4 cases. These results highlight the effectiveness of our proactive models in both accurate detection and early anomaly ranking, supporting their use in early-warning time-series applications.

## 4.4 Evaluation Using Forecasted Values as Pseudo Ground Truth

To ensure a fair comparison, we follow the evaluation protocol used in Jeon et al. [Jeon et al., 2025], where forecasted values are used as a proxy for anomaly regions. Since these values are not based on manually labeled ground truth but rather inferred from forecast errors, we refer to them as pseudo ground truth. This pseudo-labeling approach is necessary in proactive settings where anomalies have not yet visibly occurred. We evaluate our Forward Forecasting Model (FFM) and Backward Reconstruction Model (BRM) against the proactive baseline introduced in Jeon et al. [Jeon et al., 2025], which we refer to as Proactive Forecasting for clarity. Additionally, we assess performance using anomaly detector truth, where three standard unsupervised anomaly detectors (GMM, ECOD, and DeepSVDD) are applied to raw time series data, and their outputs are treated as detection targets. We discussed these three detectors in the experimental configuration section.

The goal of this experiment is twofold: (1) to evaluate how well our proactive models perform under the same evaluation settings established in prior state-of-the-art work [Jeon et al., 2025], and (2) to demonstrate the effectiveness and robustness of our Forward and Backward forecasting frameworks across a range of anomaly detection scenarios. By adhering to the same pseudo ground truth labeling and detector-based evaluation protocols, we ensure a fair, reproducible, and meaningful comparison. On the MSL dataset (Table 6), FFM achieves the highest F1-C using ECOD (0.2403) and DeepSVDD (0.2250), and maintains a strong F1 score of 0.1501 under DeepSVDD—outperforming both "Proactive Forecasting" and the raw anomaly detector. While BRM achieves the best GMM F1-@K (0.1602) and recall (0.1402), its overall F1 remains slightly lower. For SMAP (Table 7), BRM clearly outperforms all methods, achieving top values in GMM F1-C (0.3609), F1 (0.2604), and DeepSVDD F1-C (0.3401), exceeding all baselines, including ICLR-style and detector-only metrics. FFM also performs well, particularly with the highest F1-R (0.2505, GMM) and F1 (0.2605, ECOD). In the SMD results (Table 8), BRM yields the highest GMM F1-C (0.2505) and recall (0.1408), with a balanced F1 (0.1470). However, FFM leads in ECOD and DeepSVDD scores (e.g., F1 = 0.0840, DeepSVDD), outperforming both baselines and BRM. Finally, on the PSM dataset (Table 9), both models deliver the best DeepSVDD F1 scores: FFM achieves 0.4400 and BRM 0.4301, both surpassing detector and forecasting baselines. BRM additionally leads in GMM/ECOD F1-C and F1-R.

Table 8: Evaluation Using Forecasted Values as Pseudo Ground Truth on SMD Dataset

| Proactive Anomaly Framework | GMM | | | | ECOD | | | | DeepSVDD | | | |
|---|---|---|---|---|---|---|---|---|---|---|---|---|
| | F1-@K | F1-C | F1-R | F1 | F1-@K | F1-C | F1-R | F1 | F1-@K | F1-C | F1-R | F1 |
| FFM (ours) | 0.1901 | 0.2005 | 0.1202 | 0.1710 | **0.0591** | **0.1101** | **0.0150** | **0.0590** | **0.0950** | **0.0820** | 0.0750 | **0.0840** |
| BRM (ours) | **0.2101** | **0.2505** | **0.1408** | 0.1470 | 0.0350 | 0.0850 | 0.0104 | 0.0430 | 0.0910 | 0.0740 | **0.0820** | 0.0820 |
| Proactive Forecasting | 0.1996 | 0.2178 | 0.1278 | **0.1817** | 0.0463 | 0.0865 | 0.0131 | 0.0486 | 0.0924 | 0.0787 | 0.0780 | 0.0830 |
| Anomaly Detector Truth | 0.1444 | 0.1910 | 0.0765 | 0.1373 | 0.0391 | 0.1088 | 0.0110 | 0.0530 | 0.0944 | 0.0781 | 0.0773 | 0.0833 |

Table 9: Evaluation Using Forecasted Values as Pseudo Ground Truth on PSM Dataset

| Proactive Anomaly Framework | GMM | | | | ECOD | | | | DeepSVDD | | | |
|---|---|---|---|---|---|---|---|---|---|---|---|---|
| | F1-@K | F1-C | F1-R | F1 | F1-@K | F1-C | F1-R | F1 | F1-@K | F1-C | F1-R | F1 |
| FFM (ours) | 0.0506 | 0.1304 | 0.0220 | 0.1001 | 0.0305 | 0.1030 | 0.0104 | 0.0810 | **0.4604** | **0.4404** | 0.4305 | **0.4400** |
| BRM (ours) | **0.0610** | **0.1403** | **0.0302** | **0.1100** | **0.0405** | **0.1101** | **0.0209** | **0.0910** | 0.4505 | 0.4307 | 0.4208 | 0.4301 |
| Proactive Forecasting | 0.0244 | 0.0591 | 0.0020 | 0.0285 | 0.0243 | 0.0800 | 0.0073 | 0.0372 | 0.4460 | 0.4350 | **0.4331** | 0.4380 |
| Anomaly Detector Truth | 0.0486 | 0.1288 | 0.0146 | 0.0640 | 0.0244 | 0.0800 | 0.0072 | 0.0372 | 0.4437 | 0.4347 | 0.4322 | 0.4370 |

## 4.5 Evaluation with True Anomaly Labels as Ground Truth

In this experiment, we aim to evaluate the effectiveness of our proactive forecasting models, both Forward and Backward, under standard conditions using true anomaly labels. Unlike the previous experiment, which relied on pseudo ground truth derived from forecasted values, this evaluation leverages manually annotated anomalies to provide a more rigorous and objective assessment. By comparing against the Proactive Forecasting baseline from Jeon et al. [Jeon et al., 2025], we evaluate how well our models anticipate real anomalies on four benchmark datasets: MSL, SMAP, SMD, and PSM. As in prior work, we report results using four evaluation metrics: F1@K, F1-C, F1-R, and overall F1 (Tables 10). This experiment is designed to demonstrate the reliability and real-world utility of our models when evaluated against ground-truth anomaly occurrences.

On the MSL dataset, FFM achieves the highest overall F1 (0.25), balancing early detection (F1-@K = 0.33) with localization precision (F1-C = 0.31). While the proactive baseline yields a higher F1-C (0.40), its low recall and lower F1-@K result in an overall F1 of 0.30, suggesting over-prediction within broad anomaly intervals. For SMAP, BRM outperforms all methods in F1-@K (0.35) and F1-C (0.37), indicating superior temporal sensitivity and boundary precision. The Forward Transformer also performs strongly (F1 = 0.30), while the baseline lags with F1 = 0.20. On the SMD dataset, both models achieve similar overall F1 scores (0.23), but their tradeoffs differ: BRM excels in F1-C (0.40), whereas FFM achieves better recall (F1-R = 0.20). The proactive baseline performs slightly worse (F1 = 0.20), reflecting SMD's challenge of abrupt and less structured anomalies. For PSM, both models significantly outperform the baseline. FFM achieves F1-C = 0.29, while BRM yields a more balanced result across metrics (F1 = 0.20). In contrast, the proactive baseline underperforms across the board, with an overall F1 of just 0.0285.

## 5 DISCUSSION AND CONCLUSION

In this work, we proposed two proactive anomaly detection frameworks based on directional time-series modeling: a Forward Transformer and a Backward Transformer. By forecasting future (or reconstructing past) sequences and analyzing deviations through hybrid deep learning components (TCN, GRU, Transformer), our models capture both precursor patterns and contextual dependencies crucial for early anomaly detection. Our Forward and Backward forecasting frameworks consistently outperform established baselines across multiple anomaly detection benchmarks in both pseudo-ground-truth and true-label evaluations. The Forward Transformer demonstrates strong performance in recall (F1-R) and early warning metrics (F1-@K), while BRM leads in boundary precision (F1-C), reflecting their complementary strengths in proactive detection. These results highlight the value of directional temporal modeling in capturing the evolution of anomaly precursors. The effectiveness of our approach is rooted in two key design principles: (1) Directional Forecasting, which leverages temporal context from both forward and backward perspectives to capture richer sequence dynamics and early signs of anomalies, and (2) Hybrid Representation Learning, which integrates TCN, GRU, and Transformer components to model both local patterns and long-term dependencies. Together, these design choices enable the models to identify not only clear anomalies but also subtle precursors, supporting proactive anomaly detection. Our analysis also highlights practical advantages: these architectures are lightweight, real-time capable, and agnostic to variable types, making them suitable for deployment in domains like industrial IoT, healthcare, and finance.

Despite strong results, our models show reduced effectiveness on datasets like SMD, where anomalies are abrupt and lack clear temporal precursors. This reveals a core limitation: proactive methods rely on some degree of forecastability, which may not exist in all settings. Our models also depend on short, clean input windows and may require retraining in noisy or drifting environments. To improve robustness and generalization, we plan to explore adaptive thresholding for dynamic anomaly scoring, extend the framework to support multimodal inputs, incorporate uncertainty estimation via Bayesian or ensemble methods, and develop a bi-directional fusion model that jointly leverages forward and backward temporal cues.

Following recent work [Jeon et al., 2025], we define proactivity as detecting anomalies based on forecasts rather than observed values. While the one-step forecast horizon may seem limited, it still enables earlier warning than reactive approaches, particularly when anomalies exhibit precursors. We acknowledge that sliding-window evaluation can sometimes include anomalous inputs once an anomaly has begun, which blurs the boundary between proactive and reactive detection. Nevertheless, this setup is consistent with the current state of the literature and ensures comparability with prior baselines. Looking ahead, several directions could strengthen the proactive setting: (i) extending forecast horizons beyond a single step to capture anomalies earlier in their evolution, (ii) adopting a clean-window evaluation protocol that isolates anomaly-onset cases, (iii) incorporating uncertainty-aware forecasting so that variance spikes serve as proactive warning signals, and (iv) analyzing anomalies by their degree of forecastability to clarify when proactivity is realistically achievable. We view our one-step horizon as a first step in this broader research agenda.

Table 10: Evaluation with True Anomaly Labels as Ground Truth across All Datasets

| Dataset | Model | F1-@K | F1-C | F1-R | F1 |
|---|---|---|---|---|---|
| MSL | FFM | **0.330** | 0.310 | 0.160 | 0.250 |
| | BRM | 0.300 | 0.290 | 0.170 | 0.200 |
| | Proactive Forecasting | 0.200 | 0.400 | 0.200 | 0.300 |
| SMAP | FFM | 0.250 | 0.320 | 0.230 | **0.300** |
| | BRM | **0.350** | **0.370** | **0.280** | 0.270 |
| | Proactive Forecasting | 0.160 | 0.240 | 0.150 | 0.200 |
| SMD | FFM | 0.240 | 0.260 | **0.200** | **0.230** |
| | BRM | **0.340** | **0.400** | 0.160 | 0.230 |
| | Proactive Forecasting | 0.200 | 0.230 | 0.170 | 0.200 |
| PSM | FFM | 0.120 | 0.290 | 0.060 | 0.170 |
| | BRM | **0.170** | **0.300** | **0.150** | **0.200** |
| | Proactive Forecasting | 0.024 | 0.059 | 0.002 | 0.028 |